# Deep Semantic Segmentation in an AUV for Online Posidonia Oceanica Meadows Identification

Miguel Martin-Abadal, Eric Guerrero-Font, Francisco Bonin-Font and Yolanda Gonzalez-Cid
Departament de Ciencies` Matematiques` i Informatica`
Universitat de les Illes Balears
Ctra. Valldemossa Km 7.5, 07122 Palma, Spain
Email: miguel.martin, e.guerrero, francisco.bonin, yolanda.gonzalezg@uib.es

*Abstract*—Recent studies have shown evidence of a significant decline of the Posidonia oceanica (P.O.) meadows on a global scale. The monitoring and mapping of these meadows are fundamental tools for measuring their status. We present an approach based on a deep neural network to automatically perform a high-precision semantic segmentation of P.O. meadows in sea-floor images, offering several improvements over the state of the art techniques. Our network demonstrates outstanding performance over diverse test sets, reaching a precision of 96.57% and an accuracy of 96.81%, surpassing the reliability of labelling the images manually. Also, the network is implemented in an Autonomous Underwater Vehicle (AUV), performing an online P.O. segmentation, which will be used to generate real-time semantic coverage maps.

## I. INTRODUCTION

Posidonia oceanica (P.O.) is an endemic seagrass species of the Mediterranean waters that forms dense and extensive meadows, offering many benefits to the marine and coastal ecosystems [1]. Recent studies have shown evidence of a decline at alarming rates of P.O. meadows on a global scale [2] [3]. For these reasons, the European Commission directive 92/43/CEE identifies P.O. as a priority natural habitat.

A very important part of P.O. control and recovery comes through monitoring and mapping of its meadows. These are fundamental tools for measuring their status, helping to detect decline trends early on, or address the effectiveness of any protective or recovery initiative.

Nowadays, monitoring tasks are mainly carried out by divers, who measure manually meadows descriptors such as extension, shoot density or lower limit depth [4]. Nevertheless, these processes tend to be slow, imprecise and very resource-consuming.

Other approaches to monitor P.O. include the use of multi-spectral satellite imagery [5], acoustic bathymetry [6] or Autonomous Underwater Vehicles (AUV) equipped with different sensors, to extract information of P.O. meadows [7] [8]. However, these techniques suffer from lack of effectiveness in deep areas, in segregating P.O. from other algae types or are not able to perform a fully autonomous detection.

Recently, Burguera et al. [9] have achieved a fully autonomous detection by means of combining traditional image descriptors alongside Machine Learning (ML) using Support Vector Machines (SVM). Also, Gonzalez et al. [10] have explored the idea of using Convolutional Neural Networks (CNN) for P.O. detection with considerable success rates. An inconvenience of these approaches is that the classification is not made over the image as a whole, instead, the image is sub-divided into patches, which are later classified a s P.O. or background. This approach may lead to information loss, as the classification o f a p atch i s i mposed t o a ll i ts pixels.

The innovations that this work represents with respect to recent techniques in automatically identifying P.O. are: 1) the usage of a more complex deep neural network architecture that, alongside with 2) a classification b y m eans o f semantic segmentation, allows a 3) full-image pixel wise segmentation instead of a patch-based one, with no information loss or post processing needed. Finally, as a result of the aforementioned features, 4) a better accuracy is achieved in the classification task.

Our goal is to automatically perform a high-precision P.O. meadow segmentation in sea-floor i mages g athered b y a bottom-looking camera mounted on an AUV, to assess its state and evolution over time. Also, we aim to execute the neural network on an AUV, passing the segmented images to an algorithm to generate real-time semantic coverage maps of P.O. areas. These maps can be used in a dynamic path planning context to adapt the vehicle trajectory, in order to optimize the mission, in terms of duration, quality and quantity of the gathered data.

This document is structured as follows. Section II exposes the deep network architecture used and its characteristics. Following, Section III describes the different study cases, con-taining the data acquisition, processing, model tuning and val-idation process. Classification r esults a re p resented i n Section IV. Finally, Section V explains the network implementation in the AUV.

## II. DEEP LEARNING APPROACH

In the last few years, the new deep learning approaches have offered major improvements in accuracy in many computer vision tasks [11]. Causes of this are: the existence of more data, increased computation power and the development in the network architectures, making deep learning [12] one of the leading approaches in the field of computer vision.

In this work we use a semantic segmentation algorithm, based on a deep neural network, in order to achieve a segmentation of the P.O. meadows. The following subsections explain the network architecture and the training details.

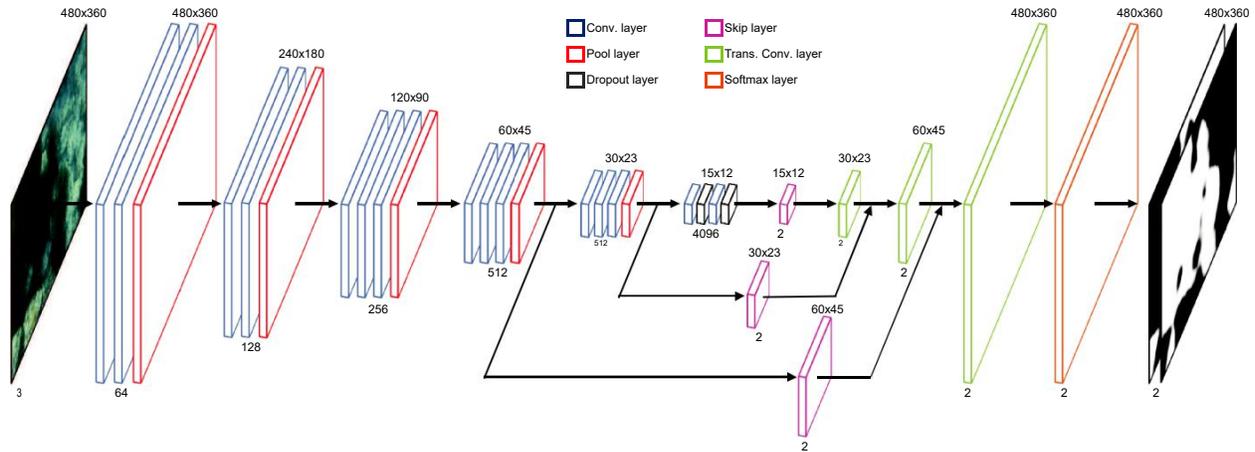

Fig. 1. Neural network architecture. Encoder: convolutional (blue), pooling (red) and dropout (black) layers. Decoder: skip (purple), transposed convolutional (green) and softmax (orange) layers. The numbers under and above the layers indicate the number of feature maps and its size, respectively.

A. Network Architecture

The architecture can be divided into two main blocks, the encoder and the decoder.

1) Encoder: The encoder purpose is to extract features and spatial information from the original images. For this task, we make use of the VGG16 architecture [13], taking out the last classification layer. This architecture uses a series of convolutional layers to extract the features, along with max pool layers to reduce the feature maps dimension. Ad-ditionally, the last two fully connected layers of the VGG16 architecture are converted into convolutional layers, in order to preserve the spatial information and obtain a first low resolution segmentation.

2) Decoder: For the decoder, we use the FCN8 architecture [14]. The decoder takes the output from the last convolutional layer of the encoder and up-samples it using transposed convo-lutional layers [15]. Also, skip layers are utilized to combine low level features from the encoder with the higher coarse information of the transposed convolutional layers. Finally, a softmax layer is applied to obtain the prediction probability for our two classes, background and P.O. The network architecture is shown in Figure 1.

This architecture, henceforth referred as VGG16-FCN8, has already presented great results in other segmentation tasks, like class segmentation of the PASCAL VOC 2011-2 dataset in [14], or road segmentation for autonomous drive in [16].

B. Training Details

The VGG16-FCN8 architecture can be trained on a single forward-backward pass. The training of the encoder is per-formed by readjusting the kernel values in the convolutional layer filters. The decoder is trained by means of the transposed convolutional and skip layer filters.

In order to train the network we need a set of images con-taining P.O., and the corresponding label map of each image, where P.O. and background areas are marked in different color codes, acting as ground truth.

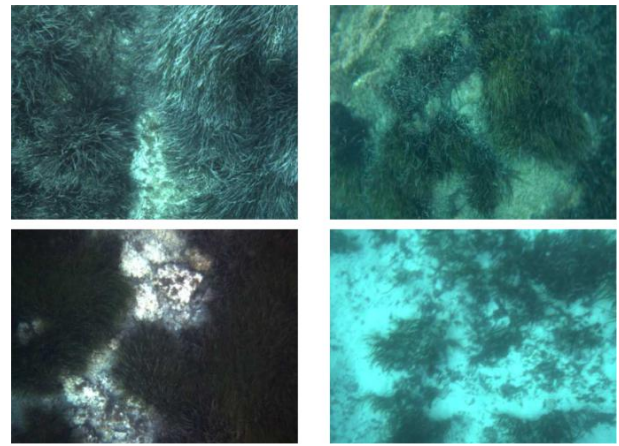

Fig. 2. Images from different missions showcasing different P.O. and water conditions.

We use a cross-entropy loss function to train the network [17], which loss increases as the predicted probability diverges from the actual label, along with the Adam optimizer [18]. Also, dropout layers with a 0.5 probability are applied to both fully connected layers of the encoder, to prevent overfitting [19].

The encoder is initialized using pretrained VGG weights on ImageNet [20]. For the decoder, the transposed convolution layers are initialized to perform bilinear upsampling. For the skip connections we apply a truncated Gaussian initialization with low standard deviation. These configuration parameters and initialization methods have already been tested, presenting great results in [16].

III. EXPERIMENTAL FRAMEWORK

This section exposes the whole experimental framework. First, it explains the acquisition and labelling of the images conforming the different datasets, along with its organization and usage. Next, the different study cases and hyperparameters used are presented. Finally, it describes the validation and evaluation details.

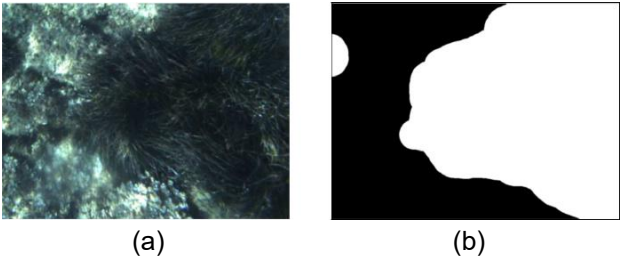

Fig. 3. (a) Original image. (b) Corresponding manually generated ground truth label map, P.O. is marked in white and background in black.

TABLE I
DATASET MANAGING

| Dataset | Location | Camera | No. Im. | Set |
|---|---|---|---|---|
| 1 | Palma Bay | Manta G283 | 164 | mix |
| 2 | Cala Blava | Manta G283 | 30 | mix |
| 3 | Valldemossa | GoPro | 157 | mix |
| 4 | Valldemossa | Manta G283 | 68 | mix |
| 5 | Valldemossa | Manta G283 | 41 | mix |
| 6 | Valldemossa | BumbleBee2 | 23 | extra |

### A. Datasets

*1) Acquisition:* The images are extracted from several video sequences obtained using three different cameras mounted alternately on the Turbot AUV: a GoPro, a stereo pair composed by two Manta G283 cameras perfectly syn-chronised and a Bumblebee2 firewire stereo rig, always facing downwards and with the lens axis perpendicular to the vehicle horizontal axis. The original image resolution is normalized and decimated to 480 360 pixels for the tests presented in this work. This reduction of the image size accelerates the segmentation process considerably, permitting its execution online. The AUV specifications and the online implementation are further developed in Section V.

Several missions were conducted on P.O. colonized coastal areas of the west and north-west of Mallorca. The objective was to obtain datasets under different P.O. conditions such as meadow density, coloration (it changes with the season and its life cycle) and health state; or water illumination, depth and turbidity, in order to build varied datasets to train and test the neural network. In all missions, the robot was programmed to move at a constant navigation altitude.

Figure 2 shows sample images from different missions showcasing different P.O. and water conditions.

*2) Labeling:* Label maps are built, manually, from the images gathered by the AUV. These label maps act as ground truth, in which the areas where P.O. is present are marked in white and the background areas in black. Figure 3 shows an original image along with its ground truth label map. It should be noticed that the boundary of the P.O. meadows is not well defined, making it hard to exactly determine the boundaries between the background and P.O. classes.

*3) Dataset Managing:* We dispose of six datasets, each one built with images extracted from video sequences recorded during the immersions, selecting sufficient images that are representative of all the aforementioned hardware and envi-ronmental conditions. We gathered one dataset from the Palma Bay, containing 164 images; another from Cala Blava, with 30 images; and four more from the Valldemossa port, of 157, 68, 41 and 23 images, respectively.

From all these datasets, two main sets of images are generated, the mix set, including 460 and the extra set with 23 images. Table I indicates the location, camera used, number of images and the corresponding set of each dataset.

The mix set (460 images) is used to train and test the network, offering a wide range of diverse and different textures containing Posidonia and thus assuring robustness in the train-ing and model selection process and also in later classification stages.

The extra set (23 images) was grabbed with a camera different from the others used to grab the videos that form the mix set, it can be used as an additional test set, allowing us to detect overfitting during the training and to assess how well the trained network generalizes on images acquired with a different camera and distinct unseen environmental conditions.

### B. Study Cases

When training a neural network, there are parameters which can be tuned, changing some of the features of the network or the training process itself. These are the so called hyperparam-eters. In order to find the values of these hyperparameters that offer the best performance, we train the network with different values and combinations, which are shown in Table II.

Firstly, we train our network with and without implementing data augmentation. Data augmentation is a technique used to reduce overfitting. It consists of applying contrast and bright-ness changes to the training images. Therefore, the network trains over more diverse data, being able to perform better on unseen conditions. On the other hand, data augmentation may cause some accuracy loss on training-like images, due to the fact that the network losses specificity during the training process [21].

Secondly, we set up two different learning rates. The learning rate value affects the size of the steps the network takes when searching for an optimal solution. Higher learning rates are able to converge more quickly, but may overshoot the optimal point. In opposition, lower learning rates converge more slowly, and may not be able to get to the optimal point [22].

Finally, we stipulate two different values for the number of iterations. This parameter sets the number of times the network backpropagates and trains. A higher number of iterations may get a better result over the training data, but also can overfit it, while fewer iterations may not be enough to reach the optimal point [22].

### C. Validation

*1) Validation Process:* We conduct eight different experi-ments, each one assessing the performance of a study case.

For each experiment, the network is trained using the corresponding study case hyperparameters. To do so, we make use of the k-fold cross validation method [23]. It consists of splitting our mix set into five equally sized subsets and train the network five times, each one using a different subset as test data and the remaining four subsets as train data. This method reduces the variability of the results, as these are less dependent on the selected test and training data, obtaining a more accurate performance estimation.

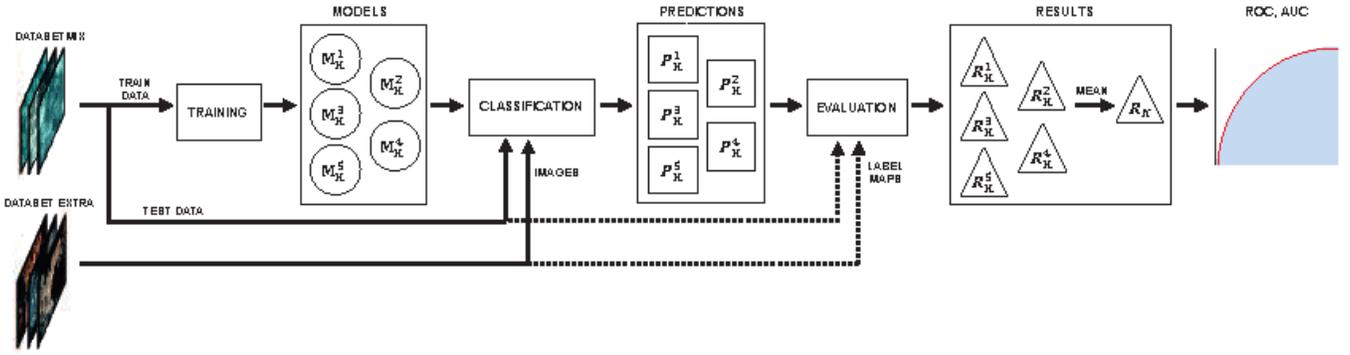

Fig. 4. Experiment "K" validation process. For each one of the eight study cases, the network is trained five times using the k-fold crossvalidation method, outputting five models. These models are run and evaluated over the mix and extra test sets. Finally, the ROC curve and AUC value are calculated from the five models mean performance.

TABLE II
STUDY CASES

| Case | Data aug. | Learning rate | Iterations |
|------|-----------|---------------|------------|
| 1 | 0 | 1e-05 | 8k |
| 2 | 0 | 1e-05 | 16k |
| 3 | 0 | 5e-04 | 8k |
| 4 | 0 | 5e-04 | 16k |
| 5 | 1 | 1e-05 | 8k |
| 6 | 1 | 1e-05 | 16k |
| 7 | 1 | 5e-04 | 8k |
| 8 | 1 | 5e-04 | 16k |

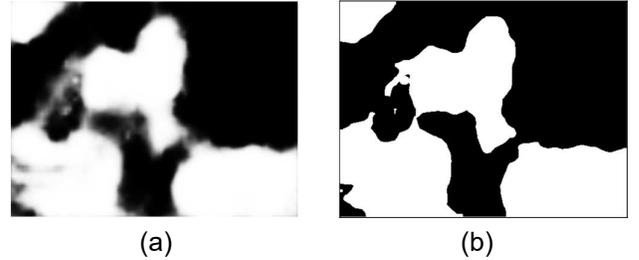

Fig. 5. (a) Probabilistic output and (b) its corresponding binarized image.

From the network training, five models are generated, $M_K^i$ where K=1..8 represents the experiment number and i=1..5 the model index. We run the five output models with their corresponding test subset and also the whole extra set, obtaining the P.O. predictions of all the models on both sets, $P_K^i$. From these predictions, each model is evaluated in order to assess its segmentation performance, $R_K^i$. The details of this process and the evaluation metrics are explained in Subsection III-C2. Finally, the segmentation performance $R_K$ of each experiment is computed as the mean of its five models performance, $R_K^i$.

From the obtained results, we generate a Receiver Operating Characteristic (ROC) curve [24]. ROC curves represent the recall against fall-out values (see equations 3 and 4) of a binary classifier at various threshold settings over the probabilistic output. We also analyse the Area Under the Curve (AUC) of the ROC curve, which gives a quantitative measure of the classifier performance. This value ranges from 0.5 to 1.0, and grows as the ROC curve is shaped to the left (low fall-out) top (high recall) corner [25].

The workflow of the validation process of the experiments is shown in Figure 4.

2) Model Evaluation Details: In order to evaluate the performance of a model, we convert the probabilistic output of the softmax layer, into a binary classification image (Figure 5). The output of the model is binarized at nine equally distributed threshold values, j=1..9.

The binarized outputs of the model are compared with the corresponding ground truth label maps. For this task, we propose a simple pixel wise comparison, analysing for each pixel if the model classification output is equal or different to its corresponding ground truth label.

From this comparison, a confusion matrix is generated, indicating the number of pixel correctly identified as P.O., True Positives (TP) and as background, True Negatives (TN); and the number of pixels wrongly identified as P.O., False Positives (FP), and as background, False Negatives (FN).

The TP, TN, FP and FN values are used to calculate the accuracy, precision, recall and fall-out of the model, defined as:

$$\text{Accuracy} = \frac{TP + TN}{TP + FP + TN + FN} \quad (1)$$

$$\text{Precision} = \frac{TP}{TP + FP} \quad (2)$$

$$\text{Recall} = \frac{TP}{TP + FN} \quad (3)$$

$$\text{Fall-out} = \frac{FP}{FP + TN} \quad (4)$$

Accuracy is defined as the percentage of correct pixel clas-sifications over all classes. Precision represents the percentage of TP classifications with respect to all the pixels classified as positives. Recall refers to the percentage of TP classifications with respect to all the truly positive pixels. Fall-out denotes the percentage of FP classifications with respect to all the truly negative pixels.

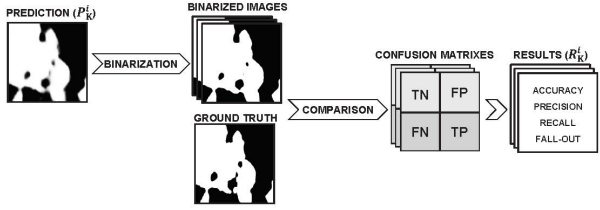

Fig. 6. Model "i" of experiment "K" evaluation process. For each model, the output prediction is binarized at j=1..9 threshold values. From every binarization "j", a confusion matrix is constructed and the accuracy, precision, recall and fall-out values are calculated.

The process followed in order to determine the segmentation performance of a model is represented in Figure 6.

## IV. CLASSIFICATION RESULTS

This section shows the results obtained for each experiment in both test sets (mix and extra), along with the hyperparam-eters selection process to build our final model. Finally, we perform a comparison of the selected model with other classi-fication methods and analyse where and why the classification errors occur.

The notation used to name each experiment makes use of three numbers. The fist one refers to the data augmentation, 0 if it is not applied, and 1 if it is. The second one indicates the learning rate value, 1 if it is 1e-05, and 5 if it is 5e-04. The third one expresses the number of iterations, 8 for 8000 and 16 for 16000. For instance, the "0 1 8" experiment refers to the experiment in which data augmentation is not applied, the learning rate is 1e-05 and the network is trained for 8000 iterations.

### A. Experiments Performance

*1) Mix set results:* First we analyse the results obtained over the test images of the mix set. Figure 7(a) represents the ROC curve along with the corresponding AUC value of each experiment. Figure 7(b) shows the precision and accuracy values obtained for each experiment at its optimal binarization threshold, selected as the one with the best (higher) trade-off between recall and fall-out, calculated as:

$$\text{Trade-off} = \frac{\text{Recall} + (1 - \text{Fall-out})}{2} \quad (5)$$

All ROC curves have an AUC over 95%, reaching a maximum of 98.7% for the 1_1_16 experiment. Following the criteria established in [26], these AUC values represent excellent classifiers.

The results show that the precision and accuracy values at optimal thresholds are greater than 90% for all the experiments. For the precision, the highest point is 96.5%, achieved in experiment 1_1_16, while the lowest one is 91.0%, obtained in experiment 0_5_8. For the accuracy, the highest point is 97.5%, achieved in experiment 1_1_8, while the lowest one is 92.2%, obtained in experiment 1_5_16.

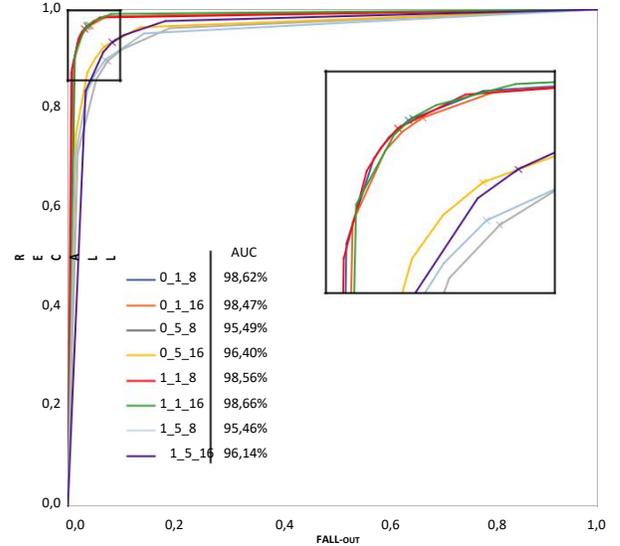

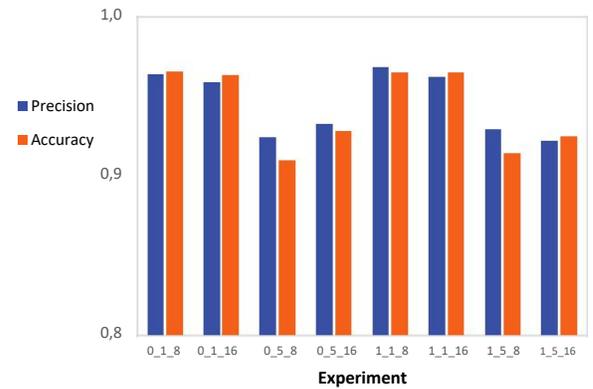

Fig. 7. Results obtained from evaluating the test images of the mix set. (a) ROC curves along with their AUC values, the optimal binarization threshold for each curve is marked with an "X". (b) Precision and accuracy values at the optimal binarization thresholds.

Experiments with the higher learning rate present slightly worse precision, accuracy and AUC values than the experiments with the lower one. On the contrary, neither the number of iterations nor the application or not of data augmentation have a significant impact on the performance.

Qualitative results of the segmentation over the mix set are shown in Figure 8.

*2) Extra set results:* While the results over the test data of the mix set are promising, as mentioned in Subsection III-A3, the test images are from the same immersions as the images used for the training and thus, the environmental conditions are similar. In order to assess the performance of the classifiers on unseen conditions, we analyse the results over the extra set, which are shown in Figure 9.

The AUC values are significantly lower for the experiments with the higher learning rate, around 92%, independently of the data augmentation state or the number of iterations. Otherwise, the experiments with the lower learning rate are able to maintain similar results as the previous test, reaching values around 97.7% when performing 16000 iterations and 97.0% when 8000. This means that these experiments do not overfit the training data, generalizing their training well enough to still perform a good classification even on images obtained with a different camera and environmental conditions that have not been trained on.

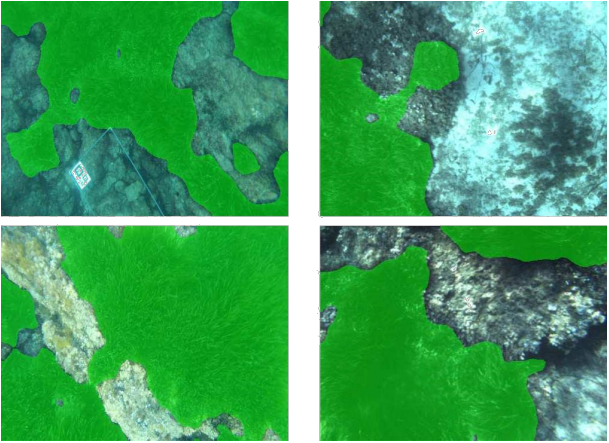

Fig. 8. Visualization of the results obtained for images from the mix set. The results of the segmentation are superimposed, in green, to the original images.

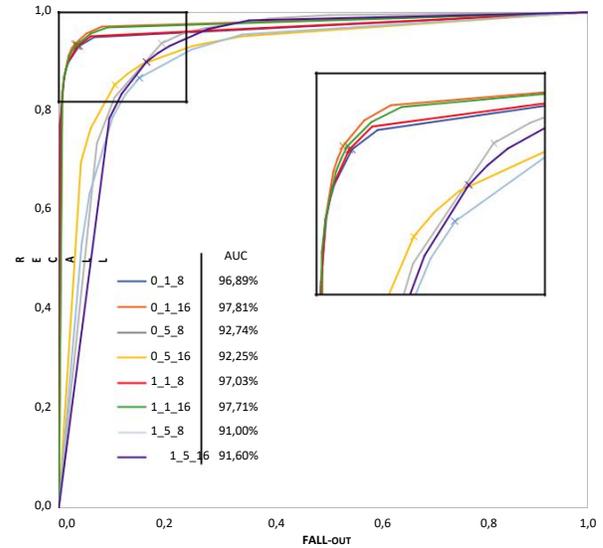

(a)

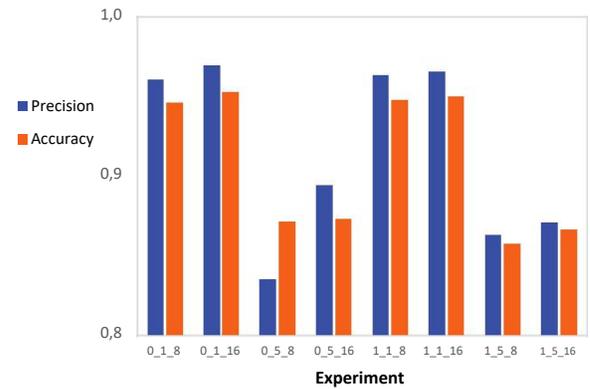

(b)

Fig. 9. Results obtained from evaluating the test images of the extra set. (a) ROC curves along with their AUC values, the optimal binarization threshold for each curve is marked with an "X". (b) Precision and accuracy values at the optimal binarization thresholds.

This can also be noticed by looking at the precision and accuracy values, calculated at the optimal binarization threshold for each experiment. The experiments with the higher learning rate achieve values around 85% for both metrics. For the experiments with the lower learning rate, the precision and accuracy values are around 96% and 95%, respectively. Again, the experiments performed with 16000 iterations have a slightly higher precision and accuracy values, while the effect of applying data augmentation or not is negligible.

Qualitative results of the segmentation over the extra set are shown in Figure 10.

### B. Hyperparameters and Model Selection

*1) Hyperparameters selection:* As a result of evaluating all experiments on both test sets, we can select the hyperparam-eters that show better performance.

Firstly, we select a learning rate of 1e-05. The results obtained on both mix and extra tests clearly show that the experiments with the lower learning rate obtain better AUC, precision and accuracy values.

Secondly, we decide to train with 16000 iterations. In the mix results we can observe that, among the lower learning rate experiments, those with a larger number of iterations have a slightly better performance.

Finally, we opt to apply data augmentation in order to generalize the training to future immersions with new unseen environmental conditions. The results show that applying it does not incur in a worse classification over the test data.

*2) Model selection:* We make an in-depth study of the performance variability for the aforementioned selected hyper-parameters by re-conducting ten times the validation process exposed in Subsection III-C, obtaining a total of fifty output models. After evaluating all models, we carry out an statistical analysis, computing the mean and standard deviation (std) of the precision and accuracy over both test sets altogether.

For the precision, the mean is 96.95% with a std of 0.97%. For the accuracy, the mean is 96.08% with a std of 0.49%. Such low std's indicate that all fifty models show a very similar performance around the mean, meaning that our network architecture and validation process are robust.

Afterwards, the model with best performance is selected from the previous fifty. This final model has a precision of 96.57% and an accuracy of 96.81%. This is the selected model to perform the online segmentation in the AUV.

*3) Comparison:* In this section we present a comparison of the VGG16-FCN8 architecture with the classification meth-ods mentioned in Section I, the Burguera et al. method [9] (henceforth ML-SVM) and the Gonzalez et al. method [10] (henceforth CNN), as well as to other state-of-the-art semantic segmentation architectures such as the U-Net [27] and the SegNet [28]. The performance comparison is conducted using the evaluation metrics defined in Section III-C2, which are obtained from the classification of the images pertaining to three test sets.

The first test set is the already known extra set, which con-tains images with new and unseen water and P.O. conditions for the classifiers.

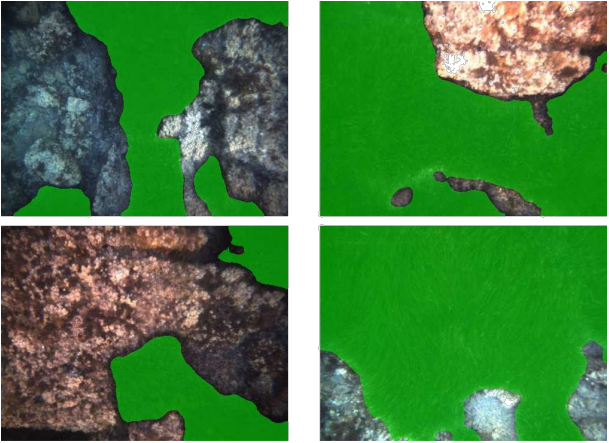

Fig. 10. Visualization of the results obtained for images from the extra set. The results of the segmentation are superimposed, in green, to the original images.

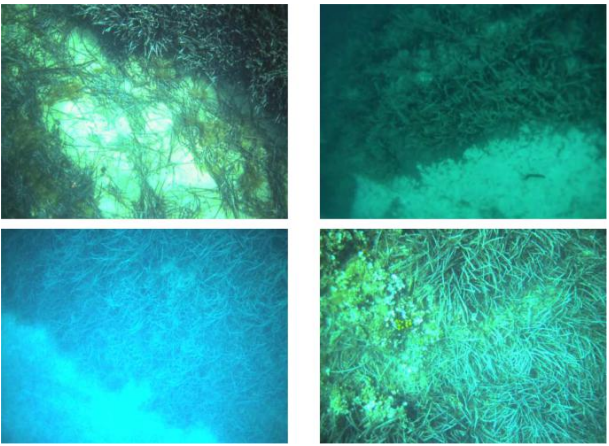

Fig. 11. Images from the croatian test set.

The second test set (henceforth, croatian set) was provided by the "Laboratory for Underwater Systems and Technologies" research group, at the University of Zagreb. It consists of 23 images extracted from video sequences recorded using a lightweight AUV by Ocenascan-MST and a Lumenera Le165 camera during different immersions in the Peljesac peninsula, Croatia. Figure 11 shows images from this test set.

Finally, the third test set (henceforth, islands set) was provided by the "Ecología Interdisciplinaria" research group, at the University of the Balearic Islands. It consists of 27 images extracted from video sequences recorded by scuba-divers using a GoPro camera during different immersions in the Mediterranean islands of Ibiza, Formentera and Menorca. Figure 12 shows images from this test set.

The croatian and islands test sets represent a challenge for the classifiers, as they were taken in new locations, following different recording procedures and using different cameras, thus, the images of these new test sets contain distinct water and P.O. conditions. Besides, the images were taken at a different distance to the P.O. meadows and with a different angle respect the sea-floor, facts that also may condition the classifiers performance.

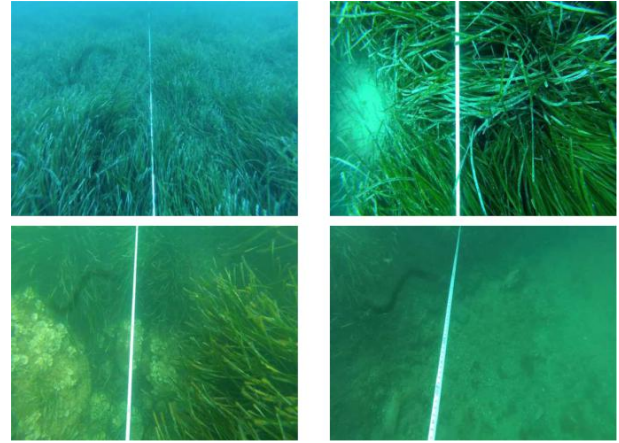

Fig. 12. Images from the islands test set.

These three sets allow us to further test the robustness of the classifiers and check their capability to be used in external applications.

For the ML-SVM method, we use the model trained over color images downsampled to 160x120 pixels and using 32x24 pixels patches, which was one of the parameter combinations that showed best results.

For the CNN method, we select the model trained using a learning rate of 1e-03 for 10 epochs with a batch size of 100.

Finally, for all semantic segmentation methods (VGG16-FCN8, U-Net and SegNet) we train them using the selected hyperparameters in Section IV-B1 and the data from the mix set.

Tables III, IV and V show the figures of the evaluation metrics of all compared classification methods over the extra, croatian and islands test sets, respectively.

We can notice that the CNN method is the worst one in all test sets, mainly due to the patch-wise classification.

The ML-SVM method seems to have been designed to be conservative when classifying the P.O. As a result, when it classifies a pixel as P.O., it is highly likely it is P.O., but the Recall and Fall-Out values denote that several pixels that truly are P.O. will be classified as background.

Consequently, it can be noticed that the ML-SVM method has a slightly better Precision than the VGG16-FCN8 when classifying the croatian and islands test sets, but the Recall and Fall-Out values are significantly worse. On the contrary, VGG16-FCN8 presents good figures in the four metrics, which implies that it is a better classifier for both P.O. and background pixels.

On the other hand, considering the three semantic seg-mentation classifiers, the U-Net and SegNet methods have a similar performance when classifying extra and croatian test sets, while U-Net shows better results when classifying the island test set. VGG16-FCN8 presents the best results of the three, suggesting again being the best semantic segmentation classifier.

To sum up, after comparing 5 different classifiers over 3 different sets of P.O. underwater images, the classifier that presents better figures in terms of the four evaluation metrics:

TABLE III
CLASSIFICATION PERFORMANCE COMPARISON OVER THE extra TEST SET

| Method | Acc. | Prec. | Recall | Fall-Out |
|---|---|---|---|---|
| ML-SVM | 89.1% | 87.1% | 94.9% | 18.0% |
| CNN | 62.2% | 81.0% | 31.9% | 7.5% |
| U-Net | 93.1% | 93.9% | 92.1% | 6.0% |
| SegNet | 90.9% | 90.4% | 91.5% | 9.7% |
| VGG16-FCN8 | 96.1% | 97.2% | 95.0% | 2.8% |

TABLE IV
CLASSIFICATION PERFORMANCE COMPARISON OVER THE croatian TEST SET

| Method | Acc. | Prec. | Recall | Fall-Out |
|---|---|---|---|---|
| ML-SVM | 66.9% | 75.0% | 37.9% | 10.0% |
| CNN | 62.0% | 79.7% | 32.1% | 8.2% |
| U-Net | 82.3% | 83.2% | 81.0% | 16.4% |
| SegNet | 83.2% | 73.5% | 82.7% | 16.3% |
| VGG16-FCN8 | 94.0% | 93.7% | 94.4% | 6.4% |

TABLE V
CLASSIFICATION PERFORMANCE COMPARISON OVER THE islands TEST SET

| Method | Acc. | Prec. | Recall | Fall-Out |
|---|---|---|---|---|
| ML-SVM | 65.7% | 88.6% | 59.5% | 19.0% |
| CNN | 67.6% | 65.7% | 73.9% | 38.6% |
| U-Net | 81.2% | 81.2% | 81.0% | 18.7% |
| SegNet | 70.3% | 70.4% | 69.8% | 29.3% |
| VGG16-FCN8 | 87.6% | 86.4% | 89.2% | 14.0% |

Precision, Accuracy, Recall and Fall-Out, is the one presented in this paper VGG16-FCN8, indicating that it is the most robust and the best option for P.O. classification in underwater images.

### C. Error Analysis

To train and evaluate the VGG16-FCN8 network we have made use of labelled images, manually generating the ground truths. This is a tedious task, subject to errors. Being aware that the evaluation of the results of the VGG16-FCN8 method could depend on the small errors present in the ground truth images, this section aims to analyse where and why the classification errors occur.

In order to do carry out this analysis we evaluate the mix set test images with the selected final model from Section IV-B2. The error analysis is conducted from the binarization of the probabilistic output at the optimal threshold.

Firstly, we perform a comparison between the binarized output and the corresponding ground truth images. The areas where these two images do not match are the FP and FN classifications. Figure 13 shows a superposition of an original image with the aforementioned comparison, marking the FN in blue and the FP classifications in green.

The majority of the errors are located on the boundaries of the P.O. meadows. As stated in Subsection III-A2, the boundary of the P.O. meadows is not well defined and hard to determine exactly, even during the manually ground truth generation process.

In order to determine if these FN and FP are really classification errors or a ground truth labelling issue, we decide to calculate the area of uncertainty of the hand labelled ground truth and see if the errors are included in it.

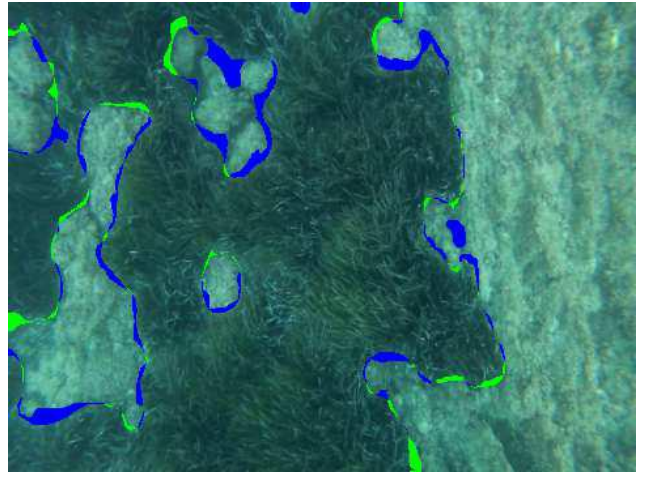

Fig. 13. Superposition of an original test image with the computed error, generated by comparing the network output with the image ground truth label map. FN are marked as blue and FP as green.

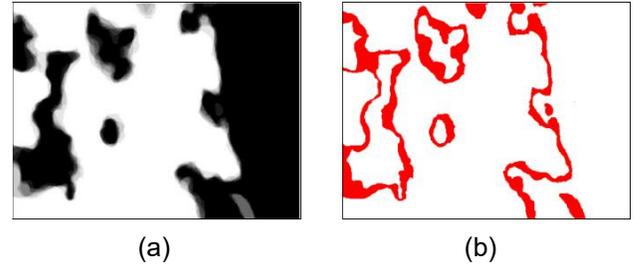

Fig. 14. (a) Mean of the manually marked label map and. (b) Area of uncertainty of the hand labelled ground truth, obtained as the area where not all ground truths match.

To do so, we ask ten people to generate the label maps of the testing images (without including the one who has generated the ground truth used to assess the network classification). Then, we compute the mean grey level for each pixel of these label maps. The areas where not all ground truth match, are marked as areas of uncertainty.

Figure 14(a) shows the computed mean label map, and 14(b) shows the obtained area of uncertainty for the original image shown in Figure 13.

For this image, a 94.6% of the misclassified pixels fall into the area of uncertainty of the hand labelled ground truth. From this, we can infer that most of the network errors do not come from misclassified pixels, but from the ground truth labelling process.

Finally, we also calculate the area of uncertainty of the neural network output as the difference in classification be-tween using 1% and 99% threshold values. This means that the uncertainty area is conformed by the pixels that the network is not entirely sure if they belong to the P.O. or background class.

Figure 15(a) shows the probabilistic output of the net when evaluating the case study image, and 15(b) shows its corresponding area of uncertainty of the neural network.

For this image, the area of uncertainty presented by the network represents an 18.9% of the whole image, while the one from the hand labelled ground truth is bigger, representing a 28.5%. As can be seen, both areas of uncertainty present a very similar shape, located on the boundaries of the P.O. meadows.

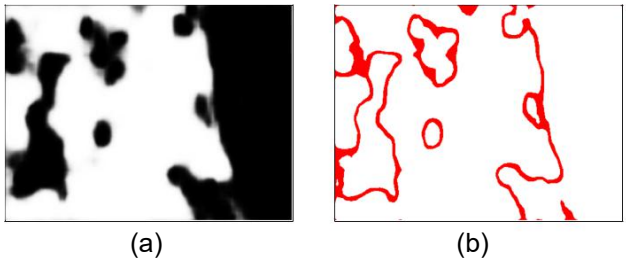

Fig. 15. (a) Probabilistic output of the network. (b) Area of uncertainty of the neural network, obtained as the classification difference when using a very high and a very low threshold.

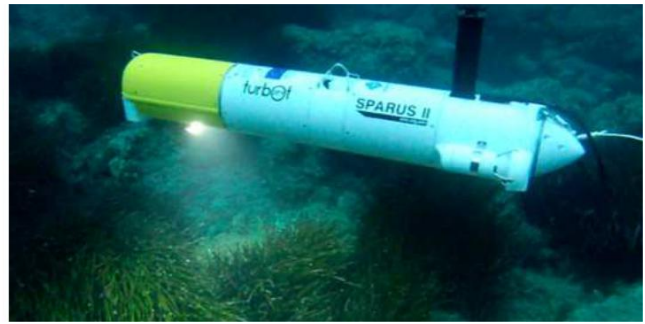

Fig. 16. Turbot AUV: SPARUS II.

These factors, along with the fact that most FN and FP are included in the uncertainty area, means that the network output is more reliable than the manually generated ground truth label map.

## V. AUV Implementation

The objective of this section is to describe the implementation of the semantic segmentation network in the AUV and its online execution, using it to generate real-time semantic coverage maps of P.O. meadows. This is carried out by sur-veying the area of interest with an AUV and recording images and their geolocalization, then, these images are processed and segmented online and passed to the coverage map generation algorithm.

In this section we present an overview of the used AUV characteristics and navigation, and the implementation of the neural network in the AUV used to perform online segmenta-tion during the robot operation.

### A. Turbot AUV

The Turbot AUV (Figure 16), property of the University of the Balearic Islands, is a SPARUS II model unit [29]. It is equipped with three motors which grant it three degrees of mobility (surge, heave and yaw). Also, it has a navigation payload, composed by: 1) a DVL (Doppler Velocity Log) to get linear and angular speeds and altitude, 2) a pressure sensor to get high frequency depth measurements, 3) an IMU (Inertial Measurement Unit) to measure accelerations and angular speeds, 4) a Compass for heading, 5) a GPS to be geo-referenced during surface navigation, and 6) an USBL (Ultra Short Baseline) acoustic link used for localization and data exchange between the robot and a remote station.

Furthermore, a stereo pair of Point Grey CM3-U3-31S4 cameras facing downwards provides the robot with images of 2048x1536 pixels resolution. These images are mainly used for three purposes: a) getting visual odometry (altitude and linear and angular speeds), b) performing online P.O. segmentation, and c) mapping the surveyed area.

The robot has two computers. One is dedicated to capturing and processing the navigation sensor data and running the main robot architecture, which is developed under the ROS middleware [30]. The second computer is where the image grabbing and online segmentation processes are executed, its specifications are: Intel i7 processor working at 2.5 GHz, 4 cores, 8GB of RAM and Ubuntu 16.04 O.S.

To perform a survey mission the vehicle must have a good estimation of its localization -Where am I?-, a well defined mission -Where should I go?-, and a proper path planning approach -How do I get there?-.

The localization of the vehicle is obtained through the fusion of multiple state estimations produced by the DVL, IMU, Compass, GPS, USBL, visual odometry and a navigation filter [31]. The survey mission is defined with a series of waypoints programmed to cover all the desired region, and with a given altitude, usually ranging between 2 and 4 meters, conditioned by the water turbidity, lighting conditions and the vehicle cruise speed. Finally, for the sake of simplicity, the strategy used by the AUV to get to the planned waypoints is a Line Of Sight (LOS) method applied to control the horizontal position using two lateral thrusters, and an altitude control using the vertical motor located at its gravity center.

### B. Online Image Segmentation

1) Implementation: To perform the online segmentation we implement a pipeline based on ROS. It loads a frozen inference graph of a trained model and executes two threads; one for the image gathering and another for the image seg-mentation.

The image gathering thread codifies every input image to RGB and then rectifies and decimates them to 480x360 pixels. The image segmentation thread receives the images and feeds them into the frozen inference graph, which generates the online P.O. segmentation.

2) Experiments: The experiments were conducted on the north coast of Mallorca, in shallow waters of 6m depth. The AUV operated at a velocity $v = 0.4 m/s$ and a navigation altitude $a = 2.5 m$.

In order to perform the segmentation of the images, it was used the frozen inference graph of the model that has shown the best performance (selected in Subsection IV-B1). The obtained segmentation framerate was $0.42\ FPS$.

An illustrative video showing the online segmentation can be seen on the SRV group web page [32]. The video shows, at the left of the screen, the video sequence captured from the camera, and at the right, the results of the segmentation superimposed in green to the original frames.

*3) Validation*: The performance is analysed in terms of the obtained framerate of the output segmentation stream. The only requirement is that, in order to avoid gaps in the generation of semantic coverage maps, the successive segmented images need to overlap.

This overlap depends on the camera displacement between two consecutive keyframes $d_{KF}$, and on the height of the image footprint $h_{FP}$. Then, the *overlap* can be expressed as:

$$overlap = (h_{FP} - d_{KF}) \cdot h_{FP}^{-1} \qquad (6)$$

where:

$$h_{FP} = (a \cdot h_{image}) \cdot f^{-1} \qquad (7)$$

$$d_{KF} = v \cdot framerate^{-1} \qquad (8)$$

Being $a$ the navigation altitude, $h_{image}$ the image height in pixels, $f$ the focal length and $v$ the AUV velocity.

Using the aforementioned vehicle speed and navigation altitude, along with an image height resolution of $h_{image} = 360$ *pixels*, a focal length of $f = 623.3$ *pixels*, and the obtained segmentation framerate. The resulting overlap is 34.0%. Thus, the framerate is high enough to get images overlap.

## VI. CONCLUSION

This section enumerates the main conclusions of this work. We have used a semantic segmentation deep network architecture to automatically perform P.O. classification in underwater images. The obtained results showed (1) very high levels of *accuracy* for diverse hyperparameters configurations, the highest one was achieved when data augmentation was applied and the network was trained with a learning rate of 1e-05 for 16000 iterations. Also, the low *std* of the evaluation metrics indicates that (2) our architecture and evaluation process are robust.

The error analysis showed that most misclassified pixels fall into the uncertainty area of the manually generated ground truth label maps. This is due to the ground truth issues caused by the fuzzy boundaries of P.O., inferring that the classification performance might be even better than the one shown on the results of the validation process.

This, along with the fact that the uncertainty area of the network is smaller than the one from the hand labelled ground truth, means that (3) the reliability of the network was higher that the manually labelling process.

Finally (4), we have implemented the segmentation process running online in an AUV operating in real environments. From the validation we obtained that the framerate of the segmented images was high enough to get images overlap, permitting an adequate semantic mapping of P.O. meadows.

Further developments will focus on lightening the online segmentation computational load while maintaining high *accuracy* levels. The aim is to provide more computational power to forthcoming autonomous exploration techniques like online mission replanning. Also, we will consider a multi-class classification, differentiating between diverse algae types and backgrounds such as rocks or sand.

The code containing the network architecture and its training process, along with the used datasets and the codes to perform the images preprocess, the output validation and the error analysis, are available on a GitHub repository [33].

## ACKNOWLEDGMENT

The authors would like to thank Antonio Vasilijevic´ extensive to all members of the Laboratory for Underwater Systems and Technologies research group at the University of Zagreb, also to A. Frank, J. Sanchez-Diaz, X. Serrano, C. Vidal, P. Ferriol and N.S.R. Agawin of the Ecolog´ıa Interdisciplinaria research group at the University of the Balearic Islands, for their invaluable contribution and support with datasets of Posidonia taken in Croacia and the Balearic Islands, key to complement the experiments presented here and to demonstrate the usefulness of the approach in several sites of the Mediterranean Sea.

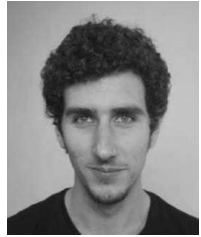

**Eric Guerrero-Font** received the B.Sc degree in mechanical engineering and the M.Sc degree in robotics and automatic control from Polytechnic University of Catalonia (BarcelonaTech). He is currently working toward the Ph.D. degree in intelligent and adaptive methods for characterization of marine environments at the University of Balearic Islands. His research activity is mainly focused on underwater robotics, machine learning and computer vision; in research topics such as intelligent control architectures and autonomous navigation.

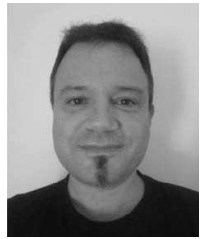

**Francisco Bonin-Font** received the degree in telecommunications engineering from the Polytechnical University of Catalonia, Barcelona, in 1996, and the Ph.D. degree in computer engineering from the University of the Balearic Islands in 2012. He has been ten years with the industry of information technology services addressed to bank business, before he initiated his academic activities. He has participated as a Technician and a Researcher in nine projects funded by the Spanish Scientific Council and the European Commission. He is also an Assistance Senior Lecturer with the Department of Mathematics and Computer Science, University of the Balearic Islands. He has authored or co-authored over 40 papers, among journals, book chapters, and conference proceedings in the field of image processing and underwater robotics, during his research activities in the Systems, Robotics and Vision Group, University of the Balearic Islands.

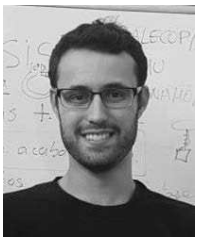

**Miguel Martin-Abadal** received the B.Sc degree in Industrial Electronics and Automation Engineering from the University of the Balearic Islands in 2015, and a M.Sc degree in Biomedical Engineering from the Polytechnic University of Madrid in 2017. He is currently doing his Ph.D. degree in Information and Communications Technology at the University of the Balearic Islands, forming part of the Systems, Robotics and Vision group. His research activity is mainly focused on computer vision, deep learning and eHealth.

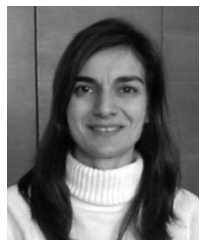

**Yolanda Gonzalez-Cid** received the Ph.D. degree in Industrial Engineering from the Spanish Open University (UNED) in 1996. She is currently an Associate Professor with the University of the Balearic Islands, Spain and member of the Systems, Robotics and Vision research group. She has been involved as participant and as leading researcher, in several projects granted by the local administration, the Spanish Scientific Council and the European Commission. Her research interests include computer vision, deep learning and eHealth.